\documentclass[10pt,journal,compsoc]{IEEEtran}
\usepackage{amsmath,amsfonts}
\usepackage{algorithmic}
\usepackage{algorithm}
\usepackage{array}
\usepackage[caption=false,font=normalsize,labelfont=sf,textfont=sf]{subfig}
\usepackage{textcomp}
\usepackage{stfloats}
\usepackage{subcaption}
\usepackage{rotating}
\usepackage[caption=false]{subfig}

\usepackage{url}
\usepackage{verbatim}
\usepackage{graphicx}
\usepackage{cite}
\usepackage{amsthm}

\usepackage{multirow}
\usepackage{booktabs}

\usepackage{pifont}
\usepackage[table]{xcolor}

\begin{document}

\title{
FedLTLib: A Comprehensive Benchmark for Federated Long-Tail Learning
}

\author{Changkun Lin, 
        Junxiao Wang
\thanks{
Corresponding author: Junxiao Wang.}
\thanks{Changkun Lin and Junxiao Wang are with Guangzhou University, China. E-mail: 2112433132@e.gzhu.edu.cn, junxiao.wang@gzhu.edu.cn}
}


\IEEEpubid{0000--0000/00\$00.00~\copyright~2021 IEEE}

\IEEEtitleabstractindextext{%
\begin{abstract}
Driven by the escalating demand for privacy-preserving computing, Federated Learning (FL) has witnessed remarkable progress, becoming a cornerstone technology for bridging distributed data silos in mobile edge networks. However, in real-world mobile computing environments, data is generated by heterogeneous mobile devices with varying user behaviors, leading to a significant Long-Tail Distribution. Unlike idealized balanced datasets, data in the wild manifests an acute imbalance where a minority of head classes dominate the sample space while a vast number of tail classes, often representing rare but critical edge-case events, are extremely scarce. This data heterogeneity, which we formally characterize as ``Double Heterogeneity'', referring to the superposition of global class imbalance and local statistical skew, precipitates severe performance deterioration on tail classes, thereby spurring the vital research direction of Federated Long-Tail Learning (FL-LT). To standardize evaluation and accelerate research in this field, we introduce FedLTLib, a comprehensive benchmark tailored for FL-LT. Addressing the critical issues of inconsistent experimental configurations and unfair comparisons in prior work, FedLTLib establishes a standardized evaluation framework. The platform not only incorporates diverse benchmark datasets reflecting mobile data characteristics but also implements 13 state-of-the-art FL algorithms (4 traditional FL algorithms and 9 FL-LT algorithms). By leveraging FedLTLib, researchers can perform fair and reproducible evaluations of algorithm robustness and generalization capabilities under a unified experimental protocol, ultimately advancing the deployment of robust intelligence in mobile computing ecosystems.
\end{abstract}

\begin{IEEEkeywords}
Federated Learning, Long-Tail Distribution, Federated Long-Tail Learning.
\end{IEEEkeywords}}
\maketitle

\section{Introduction}
\textbf{Background.}
In the era of the Internet of Things (IoT) and 5G/6G communications, the proliferation of advanced mobile devices has catalyzed the transition from cloud-based AI to mobile edge intelligence \cite{wang2025empowering}. 
Federated Learning (FL) \cite{mcmahan2017communication} has emerged as a pivotal paradigm in this shift, enabling distributed mobile nodes to collaboratively train a shared model without local data leakage. Despite its promise, the practical deployment of FL in mobile ecosystems faces a formidable challenge: the inherent long-tail distribution of ubiquitous mobile data \cite{shang2022federated}.

In mobile computing scenarios, data skewness is not merely a statistical anomaly but a fundamental reflection of user behavior heterogeneity and environmental dynamics \cite{lim2020federated}.
For instance, in mobile keyboard suggestion tasks \cite{hard2018federated}, a few common words (head classes) dominate the daily input of most users, while specialized terminology or rare expressions (tail classes) are infrequently recorded yet crucial for personalized experience.
Similarly, in mobile video surveillance and autonomous driving at the edge, common background objects are over-represented, whereas critical but rare events, such as traffic accidents or specific wildlife sightings, reside in the extreme tail of the distribution \cite{zhang2023deep}.
The geographical heterogeneity of mobile sensors often leads to a ``spatial long-tail'' effect, where certain classes are abundant in specific urban regions but nearly non-existent in others \cite{yang2021characterizing}.

Existing FL literature has extensively investigated non-IID issues \cite{li2020convergence}; however, the more pervasive problem of long-tail imbalance remains under-addressed. 
In a mobile network, tail classes often represent the edge cases that are vital for the robustness and safety of edge intelligence. Neglecting these tail classes leads to a biased model that performs poorly on rare but significant tasks, ultimately undermining the reliability of mobile services.


To address these challenges, a surge of Federated Long-Tail Learning (FL-LT) algorithms has recently emerged, diversifying into techniques such as virtual feature retraining, classifier calibration, and personalized federated learning. Historically, long-tailed recognition was predominantly matured in centralized settings, relying on strategies like class-balanced re-weighting \cite{cui2019class}, post-hoc logit adjustment \cite{menon2020long}, major-to-minor translation \cite{kim2020m2m}, or the structural decoupling of representation and classifier learning \cite{kang2019decoupling}. 
However, as the research focus pivots from centralized silos to decentralized mobile networks, the community's progress is increasingly impeded by the absence of a standardized, comprehensive, and modular evaluation benchmark.

\textbf{Motivation.} The lack of a standardized evaluation platform has led to challenges in fair comparison and reproducibility in the FL-LT domain. Specifically, current empirical efforts are plagued by the following three pain points:
\begin{itemize} 
    \item \textbf{Inconsistent Scenario Definitions.} 
    There is no consensus on the formal definition of federated long-tail distributions. As shown in Table~\ref{tab:partitioning_strategies}, many works fail to distinguish between the inherent global data imbalance and the local data heterogeneity (non-IID), leading to inconsistent experimental setups. Specifically, representative methods such as CCVR \cite{luo2021no} and FedETF \cite{2023No} operate under a globally balanced assumption, whereas specialized algorithms like FedLoGe \cite{2024fedloge} employ Pareto distributions for global modeling. When coupled with varying Imbalance Factors, these discrepancies make it hard to establish a reliable performance baseline and mask the true challenges of ``Double Heterogeneit''.

\begin{table}[t]
\centering
\caption{Comparison of Data Partitioning Strategies among Representative FL-LT Algorithms.}
\label{tab:partitioning_strategies}
\begin{tabular}{@{}lll@{}}
\toprule
\textbf{Algorithm} & \textbf{Global Partitioning} & \textbf{Local Partitioning} \\ \midrule
CReFF \cite{shang2022federated}  & Exponential Decay   & Dirichlet Partitioning \\
CCVR \cite{luo2021no}            & Global Balanced     & Dirichlet Partitioning \\
FedETF \cite{2023No}             & Global Balanced     & Dirichlet Partitioning \\
RUCR \cite{2024Federated}        & Exponential Decay   & Dirichlet Partitioning \\
CLIP2FL \cite{shi2024clip}       & VLM-Guided LT       & Label Skew Partitioning \\
FedLoGe \cite{2024fedloge}       & Pareto Distribution & Dirichlet Partitioning \\
FedIC \cite{shang2022fedic}      & Exponential Decay   & Dirichlet Partitioning \\
FedGraB \cite{xiao2023fed}       & Gradient Feedback   & Dirichlet Partitioning \\
FedYoYo \cite{yan2025you}        & Exponential Decay   & Dirichlet Partitioning \\ \bottomrule
\end{tabular}
\end{table}

    \item \textbf{Lack of Fine-grained Evaluation Metrics.} 
    A fundamental limitation in current research is the over-reliance on coarse-grained performance indicators. As shown in Table~\ref{tab:evaluation_metrics}, representative works such as CReFF~\cite{shang2022federated} and CCVR~\cite{luo2021no} primarily report overall Top-1 accuracy, which frequently obscures a model's true effectiveness on critically underrepresented tail classes. While recent studies including FedLoGe~\cite{2024fedloge} and FedYoYo~\cite{yan2025you} have begun to incorporate Many/Medium/Few accuracy, the community still lacks a systematic framework for decoupled analysis. This gap prevents a deeper understanding of whether performance gains originate from robust representation learning or more effective classifier boundary adjustment. Furthermore, most evaluations fail to account for the critical trade-off between predictive accuracy and system-level resource efficiency, such as communication and memory overhead, in resource-constrained mobile edge environments. FedLTLib addresses these deficiencies by providing a multi-dimensional evaluation suite, integrating fine-grained grouped metrics, and comprehensive system efficiency analysis.
    
    \item \textbf{Fragmented Codebases and Poor Reproducibility.} The current FL-LT landscape is characterized by highly fragmented and isolated codebases, which lack a unified software architecture. As observed in our systematic review of representative methods, researchers are frequently forced to reconcile mismatched software stacks when implementing complex procedures, such as the virtual feature distribution estimation in CCVR~\cite{luo2021no} or the fixed ETF structure initialization in FedETF~\cite{2023No}. This lack of modularity creates excessive implementation overhead and introduces inconsistent data-loading logic, which frequently compromises the reproducibility of reported state-of-the-art results. Such structural barriers significantly impede the rapid benchmarking and practical deployment of new algorithmic ideas within mobile edge intelligence ecosystems. \textit{FedLTLib} surmounts these deficiencies by providing a modular, standardized library that integrates 13 state-of-the-art algorithms, ensuring code-level transparency and facilitating seamless extensibility for the community.
\end{itemize}

\begin{table}[t]
\centering
\caption{Comparison of Evaluation Metrics among Representative FL-LT Algorithms.}
\label{tab:evaluation_metrics}
\begin{tabular}{@{}lll@{}}
\toprule
\textbf{Algorithm} & \textbf{Key Evaluation Metrics} & \textbf{Analytical Depth} \\ \midrule
CReFF~\cite{shang2022federated}  & Overall Top-1 Accuracy & Coarse-grained \\
CCVR~\cite{luo2021no}            & Overall Top-1 Accuracy & Coarse-grained \\
FedETF~\cite{2023No}             & Overall Top-1 Accuracy & Coarse-grained \\
RUCR~\cite{2024Federated}        & Overall Top-1 Accuracy & Coarse-grained \\
CLIP2FL~\cite{shi2024clip}       & Overall Top-1 Accuracy & Coarse-grained \\ \midrule
FedLoGe~\cite{2024fedloge}       & Many/Medium/Few Accuracy & Fine-grained \\
FedIC~\cite{shang2022fedic}      & Many/Medium/Few Accuracy & Fine-grained \\
FedGraB~\cite{xiao2023fed}       & Many/Medium/Few Accuracy & Fine-grained \\
FedYoYo~\cite{yan2025you}        & Many/Medium/Few Accuracy & Fine-grained \\ \bottomrule
\end{tabular}
\end{table}

\textbf{Our contribution.} 
To resolve these systemic bottlenecks and bring structural clarity to the field, we present FedLTLib, the first comprehensive, modular, and extensible benchmarking framework specifically engineered for Federated Long-Tail Learning (FL-LT). FedLTLib serves as a standardized platform mathematically designed to decouple and synthesize the intricate Double Heterogeneity, which fundamentally represents the complex superposition of global class imbalance and local statistical skew inherent in realistic mobile edge networks. By providing a unified software architecture, our framework successfully bridges the chasm between foundational federated learning pipelines and sophisticated FL-LT methodologies involving multi-stage decoupling, gradient balancing, and classifier calibration. Crucially, FedLTLib eliminates the existing engineering and evaluation barriers by integrating 13 state-of-the-art algorithms under a unified execution environment. Our primary contributions are summarized as follows:

\begin{itemize} 
    \item \textbf{A Unified Long-Tail Taxonomy.} We introduce a rigorous mathematical taxonomy that systematically decouples global class imbalance from local statistical skew, providing a formal characterization of Double Heterogeneity. Based on this formulation, we establish a standardized, parameterized data synthesis protocol capable of simulating a comprehensive spectrum of federated long-tail environments. Crucially, this protocol explicitly formalizes the highly challenging yet realistic setting where tail classes are both globally scarce and locally absent across the majority of clients, effectively eliminating existing environmental inconsistencies and laying a solid foundation for fair and reproducible benchmarking.

    \item \textbf{A Holistic and Fine-grained Evaluation Suite.} Rather than abandoning standard benchmarks, our framework establishes a multi-dimensional evaluation protocol that seamlessly pairs traditional global Top-1 accuracy with fine-grained Many/Medium/Few stratified accuracy metrics. This dual-perspective approach ensures backward compatibility with standard FL baselines while exposing localized performance variances. Furthermore, this suite incorporates hardware-centric profiling for communication payloads and empirical memory footprints, providing a holistic diagnostic toolkit tailored for resource-constrained edge intelligence.
    
   \item \textbf{A Standardized Codebase with Corrective Empirical Insights.} We open-source FedLTLib, a highly modular benchmarking library that unifies 13 state-of-the-art algorithms under a single execution framework. Through extensive systematic benchmarking, we unveil a series of critical, corrective empirical insights that expose previously unpublicized theoretical boundaries and algorithmic trade-offs:
   \begin{enumerate}
    \item \textbf{The Boundary-Representation Disconnect in Virtual Feature Retraining:} Algorithm CReFF primarily rely on synthesizing virtual features to rectify the classifier at the server side. While original studies emphasize their effectiveness in controlled settings, our benchmark reveals a severe performance bottleneck under extreme double heterogeneity. Specifically, when local clients completely lack minority samples ($\alpha \rightarrow 0$), the uploaded statistics become insufficient to capture the true underlying data distribution, inducing a severe distribution shift for the synthesized features and resulting in an unexpectedly high variance in few-shot tail recall.
    \item \textbf{The Stability of Fixed Geometric Priors in Minimalist Regimes:} FedETF fixes classifier weights as an Equiangular Tight Frame (ETF) anchored on the neural collapse phenomenon. Shifting focus away from its recognized convergence speedups, our empirical evaluations demonstrate its role as an exceptionally stable baseline for fairness-critical deployments. By completely removing the classifier from the optimization loop, it fundamentally mitigates the majority-class momentum that typically overwhelms decision boundaries, thereby sustaining more robust performance across skewed distributions than multiple sophisticated distillation-based alternatives.
    \item \textbf{The Locality-Global Conflict in Balance Augmentation:} Paradigms including FedGraB and FedLoGe introduce intricate gradient balancing or local-generic bridging mechanisms to maximize global accuracy. However, our fine-grained \textit{Many/Medium/Few} stratified analysis uncovers a critical, previously unpublicized trade-off: these methodologies frequently enhance global utility by over-correcting local optimization trajectories toward the global tail, which inadvertently compromises performance for clients whose local majority data happens to coincide with the global tail. This local utility sacrifice represents a vital fairness dimension exposed by our benchmark but largely overlooked in original evaluations.
    \item \textbf{Superiority of Self-Bootstrap and Relational Distillation:} Our empirical results demonstrate that distillation-based frameworks, specifically FedYoYo and FedIC, achieve the most distinct feature separation. By exploiting localized augmented views or calibrated prototype relationships, these algorithms maintain tight intra-class compactness for few-shot categories without requiring high-dimensional feature communication. Our t-SNE diagnostics substantiate that these techniques are highly effective at preventing feature collapse, a phenomenon where minority representations are physically subsumed by majority clusters.
\end{enumerate}

\end{itemize}

The code is available on our GitHub repository\footnote{https://github.com/gzhu-hcai/LongTailLib} for follow-up research.

\section{RELATED WORK}
For the sake of completeness, we now briefly review the Federated Learning, the corresponding statistical heterogeneity problem, existing research on long-tail distributions in the context of FL, commonly referred to as FL-LT, and the limitations of existing
benchmarks in addressing FT-LT.

\begin{figure*}[t] 
  \centering
  \makebox[\textwidth][c]{\hspace{-0.1mm}\includegraphics[width=1.1\textwidth]{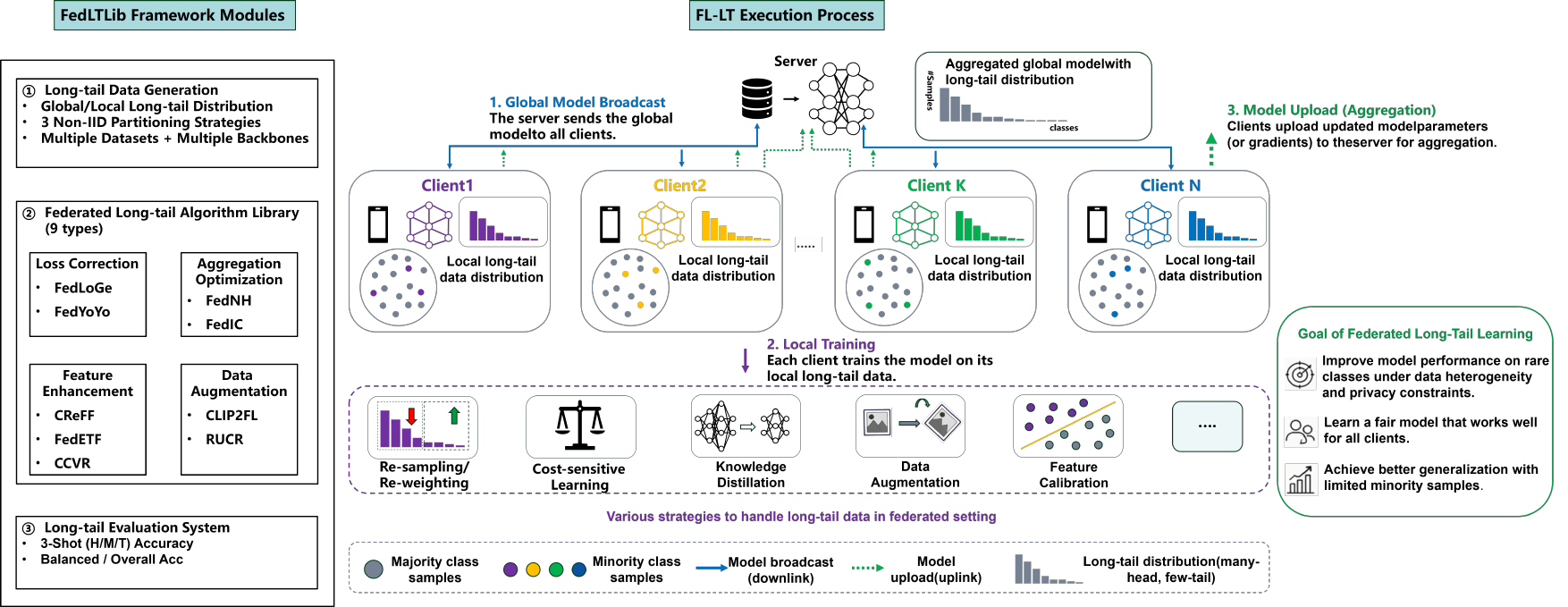}}
  \caption{The overall architecture of the FedLTLib framework, consisting of the FedLTLib Framework Modules (left) and the FL-LT Execution Process (right). The framework provides a standardized project structure for data synthesis, server-side orchestration, and client-side training. The execution process illustrates the ``Double Heterogeneity'' challenge across distinct clients and summarizes mainstream optimization paradigms integrated into our benchmark, specifically Cost-sensitive Learning, Knowledge Distillation, and Feature Calibration, to achieve global fairness and tail-class generalization.}
  \label{fig:framework}
\end{figure*}

\subsection{Federated Learning and Statistical Heterogeneity}
Federated Learning (FL) enables collaborative model training across distributed clients without compromising raw data privacy \cite{mcmahan2017communication}. A primary bottleneck in FL is statistical heterogeneity, commonly formalized as Non-IID (Independent and Identically Distributed) data \cite{hsu2019measuring}. To mitigate client drift caused by local updates on Non-IID data, standard optimization-based methods have been widely adopted. For instance, FedProx \cite{li2020federated} restricts local updates by adding a proximal regularization term to anchor local models near the global model. MOON \cite{li2021model} introduces model-level contrastive learning to maximize the agreement between local and global representations. Other prominent approaches like SCAFFOLD \cite{karimireddy2020scaffold} and FedProc \cite{mu2023fedproc} also successfully address local drift. However, these methods operate under the implicit assumption of a relatively balanced global distribution. When confronted with extreme long-tail class distributions, traditional regularization inherently favors majority classes, often leading to a catastrophic forgetting of tail classes and a collapse in global fairness.

\subsection{Federated Long-Tail Learning (FL-LT)}
FL-LT specifically targets the compounded challenge of severe class imbalance within decentralized networks \cite{zhang2024multi}. Moving beyond simple data re-sampling (which is privacy-violating in FL), recent advancements in FL-LT can be rigorously categorized into three theoretical paradigms, all of which are integrated into the FedLTLib benchmark:

\noindent\textbf{Logit Adjustment and Margin Calibration.} 
This paradigm aims to rectify the biased decision boundaries by directly calibrating the output logits or modifying the objective function during local training. FedLC \cite{zhang2022federated} calibrates the local logits based on pairwise label margins, enforcing a larger margin for tail classes to prevent head-class dominance. Similarly, methods like FedLoGe \cite{2024fedloge} dynamically adjust local optimization trajectories based on the estimated global imbalance, forcing the model to allocate more attention to minority samples without requiring explicit feature sharing.

\noindent\textbf{Virtual Feature Retraining and Prototype Alignment.} 
To circumvent the inability to access raw tail-class samples, these methods synthesize representations at the server to retrain or calibrate the classifier. CCVR \cite{luo2021no} synthesizes virtual features from Gaussian distributions parameterized by locally uploaded statistics, such as class-wise means and variances. CReFF \cite{shang2022federated} introduces learnable federated prototypes to mimic the gradient behavior of real data, allowing the server to perform decoupled classifier retraining. FEDIC \cite{shang2022fedic} further leverages the semantics and uniformity of class prototypes to guide local representation learning, significantly enhancing the feature separation for minority classes. Prototype-based sharing strategies like FedProto \cite{tan2022fedproto} also contribute to local alignment.

\noindent\textbf{Classifier Geometry and Gradient Balancing.} 
Recent theoretical insights suggest that fixing the geometry of the classifier can fundamentally eliminate the optimization bias induced by imbalanced gradients. FedETF \cite{2023No} strictly fixes the classifier weights as an Equiangular Tight Frame (ETF), forcing the network to solely optimize the feature extractor and thereby guaranteeing maximum equiangular margin. Furthermore, gradient-balancing algorithms like RUCR \cite{2024Federated} and FedGraB \cite{xiao2023fed} tackle the issue by robustly unlearning or re-weighting conflicting gradients across clients, ensuring that updates from tail-class samples are not overwhelmed by the dominant head-class momentum. To capitalize on rich pre-trained representations, CLIP-guided methods such as CLIP2FL \cite{shi2024clip} have also emerged as a powerful zero-shot calibration strategy.

\noindent\textbf{Self-Bootstrap Distillation and Representation Learning.} 
Most recently, to circumvent the reliance on external data or complex feature sharing, FedYoYo \cite{yan2025you} introduces a self-bootstrap distillation paradigm. It seamlessly couples Augmented Self-bootstrap Distillation (ASD) with Distribution-aware Logit Adjustment (DLA) to concurrently enhance minority-class representation and correct classifier biases. By treating weakly augmented local samples as self-teachers, it achieves near-centralized performance under extreme double heterogeneity without violating privacy constraints.

\begin{table*}[t]
\centering
\caption{Systematic comparison of FedLTLib with existing Federated Learning benchmarks.}
\label{tab:benchmark_comparison}
\resizebox{\textwidth}{!}{
\begin{tabular}{lccccccc}  
\toprule
\textbf{Benchmark} & \textbf{Target Problem} & \textbf{Global LT Modeling} & \textbf{Local Skew Modeling} & \textbf{Decoupled Metrics (Head/Mid/Tail)} & \textbf{Algorithm Ecosystem} \\ 
\midrule
LEAF \cite{caldas2018leaf} & General Non-IID & \ding{55} N/A & \ding{51} Empirical & \ding{55} Overall Acc. & Standard FL \\
FedML \cite{he2020fedml} & FL Systems & \ding{55} N/A & \ding{51} Dirichlet & \ding{55} Overall Acc. & Standard FL \\
FedScale \cite{lai2022fedscale} & System Efficiency & \ding{55} N/A & \ding{51} Empirical & \ding{55} Overall Acc. & System Optimizations \\
pFL-Bench \cite{chen2022pfl}& Personalized FL & \ding{55} N/A & \ding{51} Pathological/Dir. & \ding{55} Overall Acc. & pFL Methods \\
\midrule
\rowcolor{gray!15}
\textbf{FedLTLib (Ours)} & \textbf{Federated Long-Tail} & \ding{51} \textbf{Pareto/Exp. ($\rho$)} & \ding{51} \textbf{Dirichlet ($\alpha$)} & \ding{51} \textbf{Head/Mid/Tail \& Fairness} & \textbf{13 FL-LT \& SOTA Baselines} \\
\bottomrule
\end{tabular}
}
\vspace{-1em}
\end{table*}

\subsection{Benchmarks for Federated Learning}
Robust benchmarking is the cornerstone of empirical machine learning. LEAF \cite{caldas2018leaf} is the pioneering benchmark that standardizes realistic Non-IID datasets. FedML \cite{he2020fedml} provides comprehensive system-level libraries targeting topological diversity. Other specialized benchmarks focus on personalized federated learning paradigms or system efficiency.

However, as delineated in Table \ref{tab:benchmark_comparison}, existing platforms exhibit fundamental theoretical limitations when applied to FL-LT. They predominantly simulate Non-IID scenarios using a Dirichlet distribution ($Dir(\alpha)$). While this induces local label skew, it inherently maintains a relatively balanced global expectation. They lack the mathematical mechanisms to explicitly construct and explicitly decouple Global Long-Tailness, typically modeled via Pareto decay distributions with specific Imbalance Factors, from Local Heterogeneity. Moreover, relying solely on overall accuracy actively conceals model degradation on tail categories. FedLTLib is precisely engineered to fill this critical void, providing the first mathematically rigorous data synthesis pipeline for double long-tail scenarios and a standardized evaluation protocol for decoupled algorithms.

\section{Problem Formulation and Taxonomy}

\subsection{Standard Federated Classification}
We consider a standard federated image classification task with $M$ distributed clients. Let $\mathcal{X}$ denote the input feature space and $\mathcal{Y} = \{0, 1, \dots, K-1\}$ denote the label space with $K$ distinct classes. The $i$-th client possesses a local dataset $\mathcal{D}_i = \{(\mathbf{x}_{i,j}, y_{i,j})\}_{j=1}^{|\mathcal{D}_i|}$ drawn from a local distribution $\mathcal{P}_i(\mathcal{X}, \mathcal{Y})$. The overarching objective of FL is to collaboratively learn a global model $f_{\mathbf{w}}: \mathcal{X} \rightarrow \mathbb{R}^K$ parameterized by $\mathbf{w}$, which minimizes the global empirical risk without centralizing the local datasets:
\begin{equation}
\min_{\mathbf{w}} \mathcal{L}(\mathbf{w}) = \sum_{i=1}^M \frac{|\mathcal{D}_i|}{|\mathcal{D}|} \mathcal{L}_i(\mathbf{w})
\label{eq:fl_objective}
\end{equation}
where $|\mathcal{D}| = \sum_i |\mathcal{D}_i|$ is the total number of samples across all clients, and $\mathcal{L}_i(\mathbf{w})$ is the local objective function typically defined by the Cross-Entropy loss.

\subsection{Mathematical Formulation of Double Heterogeneity}
In realistic FL-LT scenarios, the data generation process is governed by a double-heterogeneity model: the Global Class Imbalance and the Local Statistical Skew.

\noindent\textbf{Definition 1: Global Long-Tailness.} 
Let $N_k$ denote the total number of samples for class $k \in \mathcal{Y}$ across the entire federated network, i.e., $N_k = \sum_{i=1}^M N_{k,i}$, where $N_{k,i}$ is the number of class $k$ samples on client $i$. Without loss of generality, we assume the classes are sorted in descending order of their global frequencies, such that $N_0 \geq N_1 \geq \dots \geq N_{K-1}$. The macroscopic imbalance is quantified by the Imbalance Factor ($IF$), defined as the ratio of the majority (head) class size to the minority (tail) class size:
\begin{equation}
IF = \frac{N_0}{N_{K-1}}
\label{eq:imbalance_factor}
\end{equation}
To explicitly model the continuous long-tail decay observed in real-world datasets, we parameterize the global distribution using an exponential decay function \cite{zhang2023deep}. The sample size for intermediate classes is deterministically generated by a decay factor $\rho \in (0, 1]$:
\begin{equation}
N_k = N_0 \cdot \rho^k
\label{eq:exponential_decay}
\end{equation}
Consequently, the macroscopic $IF$ and the microscopic decay factor $\rho$ are mathematically equivalent and firmly coupled via the formulation:
\begin{equation}
IF = \frac{N_0}{N_0 \cdot \rho^{K-1}} = \left(\frac{1}{\rho}\right)^{K-1} \iff \rho = \left(IF\right)^{-\frac{1}{K-1}}
\label{eq:rho_if_mapping}
\end{equation}
This explicit mapping allows \textit{FedLTLib} to precisely synthesize arbitrary degrees of global imbalance by simply adjusting the $IF$.

\noindent\textbf{Definition 2: Local Label Skew via Dirichlet Partitioning.} 
Given the globally synthesized samples $N_k$ for class $k$, these samples must be allocated to $M$ clients to simulate localized Non-IID constraints. We employ a Dirichlet distribution \cite{hsu2019measuring} to govern this allocation. Specifically, the proportion of class $k$ samples allocated to client $i$, denoted as $q_{k,i}$, is sampled from:
\begin{equation}
\boldsymbol{q}_k = [q_{k,1}, q_{k,2}, \dots, q_{k,M}] \sim \text{Dir}(\alpha \mathbf{1}_M)
\label{eq:dirichlet_partition}
\end{equation}
where $\alpha > 0$ is the concentration parameter. The actual sample size of class $k$ on client $i$ is exactly $N_{k,i} = \lfloor N_k \cdot q_{k,i} \rfloor$. A smaller $\alpha$ induces extreme local label skew (some clients may entirely lack certain classes), whereas $\alpha \rightarrow \infty$ recovers an IID partition where $P_i(y) \approx P_{global}(y)$.

\subsection{The Unified FL-LT Taxonomy}
By mathematically decoupling the global imbalance factor ($IF$, as defined in Eq. \ref{eq:imbalance_factor}) and the local heterogeneity parameter ($\alpha$, as defined in Eq. \ref{eq:dirichlet_partition}), FedLTLib establishes the first unified taxonomy for Federated Long-Tail Learning. We categorize the complex evaluation space into three primary scenarios:

\begin{itemize}
    \item \textbf{Scenario A: Global Balance + Local Skew ($IF=1, \alpha \to 0$).} This represents the traditional Non-IID FL research paradigm. The global dataset is perfectly balanced, but local clients observe highly skewed sub-distributions.
    \item \textbf{Scenario B: Global Imbalance + Local IID ($IF \gg 1, \alpha \to \infty$).} This mirrors the conventional centralized long-tail learning paradigm transposed to a distributed setting. Local clients share identical long-tail distributions.
    \item \textbf{Scenario C: Double Long-Tail ($IF \gg 1, \alpha \to 0$).} The most challenging and realistic setting. The global distribution is intrinsically long-tailed, and the local data is concurrently heavily skewed. In this scenario, tail classes are not only globally scarce but also entirely absent from the majority of clients, inducing severe gradient conflicts and catastrophic forgetting.
\end{itemize}

\section{Design of the FedLTLib Framework}

Our proposed framework, FedLTLib, is designed to systematically evaluate and advance research in Federated Long-Tail Learning (FL-LT). The framework is designed to decouple data distribution simulation, federated communication mechanisms, and long-tail optimization algorithms, aiming to provide researchers with a standardized and highly reproducible experimental platform.

\subsection{System Architecture and Module Decoupling}

The core architecture of FedLTLib consists of three independent yet highly collaborative modules: the Data Partitioner, the Federated Orchestration Engine, and the Algorithm Zoo.

In the federated orchestration engine, we adopt a standard Client-Server communication topology. In the $t$-th communication round, the central server broadcasts its global model parameters $w^t$ to the selected set of clients $S_t$. Each client $k \in S_t$ performs local training for $E$ epochs based on its local long-tail dataset $D_k$, with the optimization objective:
\begin{equation}
\min_{w} F_k(w) = \mathbb{E}_{(x,y) \sim \mathcal{D}_k} [\mathcal{L}(f(x; w), y)]
\end{equation}
where $\mathcal{L}(\cdot)$ is the loss function, and $\mathcal{D}_k$ represents the local data distribution of the $k$-th client. Upon completing the local updates, the client uploads the model increment $\Delta w_k^t$ to the server for aggregation:
\begin{equation}
w^{t+1} = w^t + \eta_g \sum_{k \in S_t} \frac{|D_k|}{\sum_{j \in S_t} |D_j|} \Delta w_k^t
\end{equation}
where $\eta_g$ is the global learning rate at the server side. Through a unified two-layer abstract interface, this decoupled design enables researchers to seamlessly plug in customized algorithms, ranging from Personalized FL to Adversarial Training strategies, without modifying the underlying communication logic or data loading modules.

\subsection{Benchmark Datasets and Double Distribution Modeling}

Standardized and unified data construction is a fundamental prerequisite for fair evaluation. FedLTLib natively supports three classic benchmark datasets with increasing resolutions and categorical complexities: CIFAR-10 \cite{krizhevsky2009learning}, CIFAR-100 \cite{krizhevsky2009learning}, and Tiny-ImageNet \cite{le2015tiny}.

Unlike traditional centralized long-tail learning, data imbalance in federated scenarios manifests as a complex ``double heterogeneity'': the superposition of global label distribution skew and local non-IID (Independent and Identically Distributed) characteristics. FedLTLib introduces a parameterized distribution generator to mathematically formulate this complexity across the supported datasets.

Suppose the dataset contains $C$ classes. For the global long-tail distribution, we employ an exponential decay model to control the decreasing trend of the sample size per class. The total number of samples $N_c$ for class $c \in \{1, 2, \dots, C\}$ is defined as:
\begin{equation}
N_c = N_1 \cdot \mu^{c-1}
\end{equation}
where $\mu \in (0, 1)$ is the decay coefficient. The Imbalance Factor (IF), denoted as $\rho_{global}$, is strictly defined as the ratio of the maximum class sample size to the minimum class sample size, i.e., $\rho_{global} = \frac{N_1}{N_C}$.

At the level of local non-IID partitioning, we utilize a sampling strategy based on the Dirichlet distribution to map the global data to $K$ clients. For class $c$, its sample proportion vector allocated to client $k$, denoted as $\mathbf{p}_c$, follows $\mathbf{p}_c \sim \text{Dir}(\alpha \mathbf{1}_K)$. The concentration parameter $\alpha > 0$ directly controls the skewness of the local data distribution: as $\alpha \to 0$, clients contain samples from only a very few classes (extreme non-IID); as $\alpha \to \infty$, the local distribution degenerates to an IID partition consistent with the global long-tail distribution.

\subsection{Comprehensive Algorithm Zoo}

To establish objective performance baselines and cover mainstream technical paradigms, FedLTLib integrates 13 representative algorithms (4 traditional FL baselines and 9 specialized FL-LT algorithms). As systematically summarized in Table \ref{tab:algorithm_zoo}, we construct a comprehensive taxonomy of these methods based on their optimization objectives, core mechanisms, and privacy dependencies:

\begin{itemize}
    \item \textbf{Model Component Improvement:} This category encompasses Virtual Feature Retraining techniques, including CReFF [3]\footnote{\url{https://github.com/shangxinyi/CReFF-FL}}, CLIP2FL [27]\footnote{\url{https://github.com/shijiangming1/CLIP2FL}}, and CCVR [21]\footnote{\url{https://github.com/smduan/Fed-CCVR}}, which generate virtual samples via cross-client feature aggregation without compromising privacy. It also covers Classifier Calibration methods, such as RUCR [25]\footnote{\url{https://github.com/liuyuxia211/RUCR}} and FedETF [24]\footnote{\url{https://github.com/ZexiLee/ICCV-2023-FedETF}}, designed to adjust decision boundaries and alleviate head-class dominance.

    \item \textbf{Algorithm-Based Optimization:} This paradigm involves Knowledge Distillation methods like FedIC [22]\footnote{\url{https://github.com/shangxinyi/FEDIC}} and FedYoYo [28]\footnote{\url{https://github.com/shanss132/FedYoYo}}, alongside Balance Augmentation approaches such as FedGraB [26]\footnote{\url{https://github.com/ZackZikaiXiao/FedGraB}} and FedLoGe [20]\footnote{\url{https://github.com/ZackZikaiXiao/FedLoGe}}, which dynamically adjust gradient updates or representation learning based on global long-tailed distribution feedback. Notably, the recently integrated FedYoYo introduces an Augmented Self-bootstrap Distillation (ASD) paradigm to enhance minority-class representation entirely locally, avoiding the need for external data sharing.
\end{itemize}

Furthermore, at the loss function level, FedLTLib naturally incorporates Logit Adjustment \cite{menon2020long}, which serves as a foundational component for margin-based methods like FedLC and the Distribution-aware Logit Adjustment (DLA) in FedYoYo. For a sample with class frequency $\pi_c$, we introduce a posterior-adjusted Softmax cross-entropy during local training:
\begin{equation}
\mathcal{L}_{LA}(y, f(x)) = -\log \frac{\exp(f_y(x) + \tau \log \pi_y)}{\sum_{c=1}^C \exp(f_c(x) + \tau \log \pi_c)}
\end{equation}
where $\tau$ is a hyperparameter controlling the long-tail smoothness.

\begin{table*}[htbp]
  \centering
  \caption{Summary of the 13 algorithms integrated into the FedLTLib Algorithm Zoo.}
  \label{tab:algorithm_zoo}
  \resizebox{0.95\textwidth}{!}{
  \begin{tabular}{llcll}
    \toprule
    \textbf{Category} & \textbf{Algorithm} & \textbf{Venue \& Year} & \textbf{Core Mechanism / Optimization Strategy} & \textbf{Privacy Dependency} \\
    \midrule
    \textbf{Traditional Baselines} 
    & FedAvg \cite{mcmahan2017communication} & AISTATS 2017 & Vanilla parameter averaging & None \\
    & FedProx \cite{li2020federated} & MLSys 2020 & Proximal term regularization for local drift & None \\
    & MOON \cite{li2021model} & CVPR 2021 & Model-level contrastive learning & None \\
    & FedLC \cite{zhang2022federated} & ICML 2022 & Logit calibration based on label distribution & Label stats \\
    \midrule
    \textbf{FL-LT Algorithms} 
    & CReFF \cite{shang2022federated} & IJCAI 2022 & Virtual feature retraining via federated prototypes & Prototypes \\
    & CCVR \cite{luo2021no} & NeurIPS 2021 & Classifier calibration with virtual representations & Feature stats \\
    & FedETF \cite{2023No} & IJCAI 2023 & Fixed Equiangular Tight Frame (ETF) classifier & None \\
    & RUCR \cite{2024Federated} & IEEE TIFS 2024 & Representation unification \& classifier rectification & None \\
    & CLIP2FL \cite{shi2024clip} & AAAI 2024 & Zero-shot calibration guided by CLIP features & External VLM \\
    & FedLoGe \cite{2024fedloge} & ICLR 2024 & Joint local and generic optimization bridging & None \\
    & FedIC \cite{shang2022fedic} & ICME 2022 & Calibrated distillation for feature separation & Logits / Protos \\
    & FedGraB \cite{xiao2023fed} & NeurIPS 2023 & Self-adjusting gradient balancer & Gradient stats \\
    & FedYoYo \cite{yan2025you} & ICCV 2025 & Self-bootstrap distillation \& distribution-aware logit adjustment & Label stats \\
    \bottomrule
  \end{tabular}
  }
\end{table*}

\subsection{Multi-dimensional Evaluation Protocol}

FedLTLib features a rigorous, multi-dimensional evaluation protocol that comprehensively assesses both algorithmic prowess and engineering practicality:

\textbf{Generalization Performance:} To address the evaluation bias caused by global Top-1 Accuracy in long-tail scenarios, we introduce stratified Recall and F1-Score calculations based on Many-shot ($N_c > 100$), Medium-shot ($20 < N_c \leq 100$), and Few-shot ($N_c \leq 20$) categories. This fine-grained quantification accurately reflects the model's generalization capability on tail data. Additionally, client-level local accuracy variance is monitored to assess algorithmic fairness.

\textbf{System Efficiency Evaluation:} Recognizing the strict constraints of edge deployments, FedLTLib systematically tracks operational overheads. Key metrics include Communication Overhead(total bytes uploaded and downloaded across communication rounds), Memory Footprint(peak RAM/VRAM usage on both server and client sides), and Convergence Speed(global epochs required to attain a predefined accuracy threshold).

\subsection{Plug-and-Play Extensibility Interface}
Beyond the pre-configured benchmark datasets, FedLTLib is engineered with high extensibility in mind. We abstract the complex double-heterogeneity sampling logic into a unified, plug-and-play data interface, implemented as a standardized DataPartitioner module. Researchers aiming to evaluate algorithms on novel domain-specific datasets, such as medical imaging or autonomous driving logs, merely need to inherit our standard dataset API. The framework automatically applies the parameterized global long-tail decay and local Dirichlet partitioning, completely decoupling the raw data ingestion from the statistical heterogeneity simulation. This modular design eliminates prohibitive engineering overhead and ensures the framework's long-term vitality in the FL community.

\section{Experiments and Evaluation}
To comprehensively evaluate the robustness and generalization capabilities of various federated learning algorithms under double heterogeneity, we conducted extensive experiments using the proposed FedLTLib benchmark. Drawing inspiration from standard evaluation methodologies, our experimental design and subsequent analyses aim to answer the following four core research questions (RQs):

\begin{itemize}
    \item \textbf{RQ1: (Robustness to Global Imbalance).} Compared to traditional FL baselines, how effectively do specialized FL-LT algorithms maintain global predictive performance across varying degrees of long-tail severity (Imbalance Factors)?
    
    \item \textbf{RQ2: (Fine-grained Generalization on Tail Classes).} Beyond coarse-grained overall accuracy, can specialized interventions effectively prevent the catastrophic sacrifice of tail classes and ensure fair classification performance across Head, Medium, and Tail categories?
    
    \item \textbf{RQ3: (Optimization Stability and Feature Separability).} How do different optimization paradigms influence the federated learning dynamics, and can they prevent minority classes from collapsing into the feature clusters of majority classes?
    
    \item \textbf{RQ4: (System Efficiency and Resource Trade-offs).} What are the practical overheads of these state-of-the-art algorithms regarding theoretical communication payload and peak memory footprint, and are they feasible for deployment on resource-constrained edge devices?
\end{itemize}

The following sections are structured to address these RQs sequentially, beginning with the experimental setup and double heterogeneity visualization.

\subsection{Experimental Setup and Double Heterogeneity Visualization}

\begin{table}[htbp]
\centering
\caption{Statistical Summary of the Benchmark Datasets}
\label{tab:datasets}
\resizebox{\columnwidth}{!}{%
\begin{tabular}{lcccc}
\toprule
\textbf{Dataset} & \textbf{Resolution} & \textbf{Classes} & \textbf{Training Samples} & \textbf{Test Samples} \\
\midrule
CIFAR-10         & $32 \times 32$      & 10               & 50,000                    & 10,000                \\
CIFAR-100        & $32 \times 32$      & 100              & 50,000                    & 10,000                \\
Tiny-ImageNet    & $64 \times 64$      & 200              & 100,000                   & 10,000                \\
\bottomrule
\end{tabular}%
}
\end{table}

\begin{figure*}[t]
    \centering
    \subfloat[Global Imbalance ($IF=50$)\label{fig:global_imbalance}]{
        \includegraphics[width=0.48\linewidth]{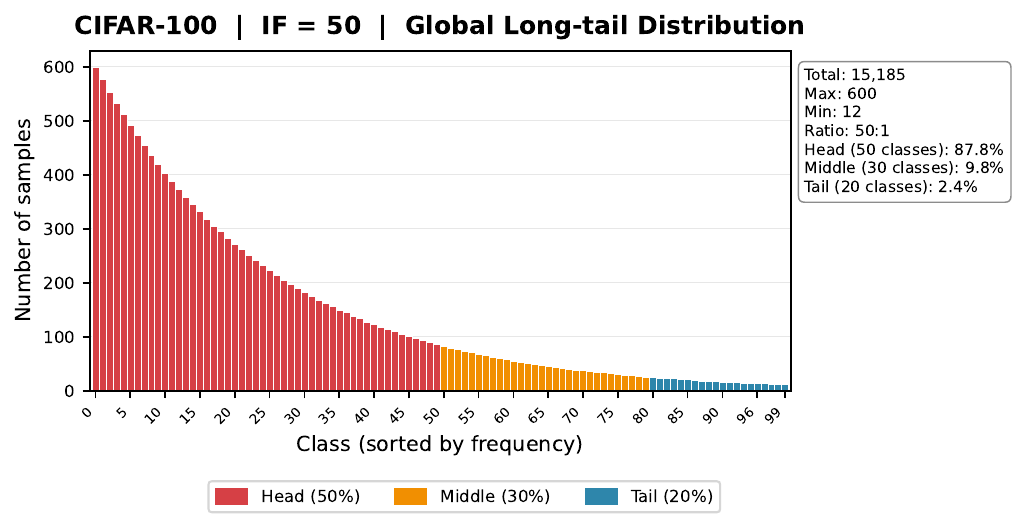}
    }\hfill
    \subfloat[Local Statistical Skew ($\alpha=0.5$)\label{fig:local_skew}]{
        \includegraphics[width=0.48\linewidth]{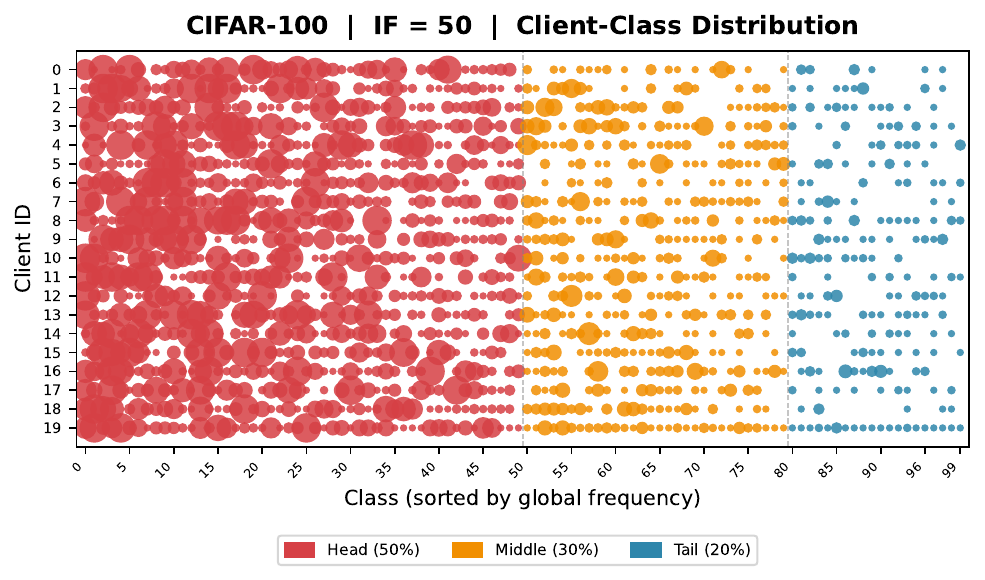}
    }
    
    \vspace{-2mm}
    \caption{Visualization of the proposed Double Heterogeneity under the rigorous Scenario C configuration (using CIFAR-100 at $IF=50$ and $\alpha=0.5$ as a representative prototype). (a) Global Imbalance: The aggregated global distribution exhibits a steep exponential decay, severely starving the Tail classes. (b) Local Statistical Skew: The bubble chart illustrates the extreme non-IID data partition across 20 simulated edge clients. The severe fragmentation and extensive blank regions in the right hemisphere visually demonstrate that tail categories are completely absent across the vast majority of local clients.}
    \label{fig:double_heterogeneity}
\end{figure*}

To systematically evaluate the efficacy and scalability of existing algorithms under the proposed FedLTLib framework, we conduct extensive benchmark analyses across three image classification datasets of increasing resolution and categorical complexity: CIFAR-10, CIFAR-100, and Tiny-ImageNet. The fundamental statistical characteristics of these datasets are formally summarized in TABLE \ref{tab:datasets}.

Before delving into the multi-dimensional quantitative metrics, we utilize CIFAR-100 as a highly representative prototype to visually demonstrate the core challenge of our benchmark, termed Double Heterogeneity, in Figure~2. Across our empirical evaluations, the system is fundamentally configured to simulate the highly challenging Scenario C (Double Long-Tail). As visualized in Figure \ref{fig:double_heterogeneity}, we employ a severe global Imbalance Factor (IF) of 50 paired with a local Dirichlet concentration parameter $\alpha=0.5$. The aggregated global distribution exhibits a steep exponential decay, severely starving the tail classes of representative samples. Compounding this challenge, the corresponding bubble chart illustrates extreme local non-IID data partitions across 20 simulated edge clients. The severe fragmentation and extensive blank regions visually expose the complete absence of tail classes across the vast majority of individual clients. 

Crucially, this rigorous setting realistically simulates an increasingly severe federated environment. When transitioning from the 10 classes of CIFAR-10 to the denser label spaces of CIFAR-100 and the 200 classes of Tiny-ImageNet, the collision of global scarcity and extreme local absence is drastically amplified. This forms the exact Double Heterogeneity dilemma. We utilize this comprehensive testbed to systematically evaluate the generalization and robustness of 4 traditional FL baselines and 9 specialized FL-LT algorithms across diverse complexity scales.

\begin{table*}[htbp]
  \centering
  \small
  \setlength{\tabcolsep}{15pt} 
  \caption{Global Test Accuracy (\%) across different Imbalance Factors (IF) on CIFAR-10/100 with $\alpha=0.5$.}
  \label{tab:if_robustness}
  \begin{tabular}{llcccccc}
    \toprule
    \multirow{2}{*}{Category} & \multirow{2}{*}{Method} & \multicolumn{3}{c}{CIFAR10-LT} & \multicolumn{3}{c}{CIFAR100-LT} \\
    \cmidrule(lr){3-5} \cmidrule(lr){6-8}
    & & IF=100 & IF=50 & IF=10 & IF=100 & IF=50 & IF=10 \\
    \midrule
    \multirow{4}{*}{\shortstack[l]{Traditional \\ Baselines}}
    & FedAvg  & 64.34 & 72.20 & 84.45 & 40.60 & 46.33 & 61.12 \\
    & FedProx & 61.92 & 66.50 & 81.06 & 37.93 & 42.71 & 57.50 \\
    & FedLC   & 66.26 & 71.86 & 84.82 & 40.32 & 46.02 & 60.87 \\
    & MOON    & 64.99 & 71.29 & 82.82 & 40.57 & 45.68 & 60.61 \\
    \midrule
    \multirow{9}{*}{\shortstack[l]{Specialized \\ FL-LT Algorithms}}
    & CReFF   & 63.09 & 64.63 & 78.43 & 34.09 & 37.54 & 48.50 \\
    & CLIP2FL & 71.66 & 73.40 & 82.42 & 38.16 & 42.29 & 54.19 \\
    & CCVR    & 71.20 & 75.30 & 85.56 & 42.87 & 47.46 & 62.46 \\
    & RUCR    & 65.14 & 72.74 & 84.68 & 39.24 & 44.33 & 58.70 \\
    & FedETF  & 69.67 & 74.12 & 84.25 & 35.29 & 42.07 & 54.95 \\
    & FedLoGe & 71.65 & 80.23 & 89.85 & 44.99 & 51.06 & 67.77 \\
    & FedIC   & 71.74 & 71.99 & 73.79 & 31.25 & 34.46 & 40.69 \\
    & FedGraB & 68.82 & 78.15 & 88.38 & 41.85 & 49.08 & 65.88 \\
    & FedYoYo & 80.54 & 82.99 & 88.70 & 52.82 & 57.65 & 69.84 \\
    \bottomrule
  \end{tabular}
\end{table*}

\subsection{Robustness to Global Imbalance (Overall Performance)}

To comprehensively assess algorithmic robustness against varying degrees of global long-tailed skewness and categorical complexity, we extract the best global Top-1 accuracy across three Imbalance Factors ($IF \in \{10, 50, 100\}$) on both the CIFAR-10 and CIFAR-100 datasets, as systematically summarized in Table~5. 

The quantitative results unveil several critical insights regarding the scaling limits of current algorithms under extreme double heterogeneity:

\noindent\textbf{1) Universal Degradation and Accelerated Baseline Collapse.} 
While all methodologies experience performance degradation as the Imbalance Factor escalates, traditional FL baselines exhibit an accelerated collapse when categorical complexity increases. For instance, FedAvg and FedProx achieve 64.34\% and 61.92\% respectively on CIFAR-10 at $IF=100$, yet they precipitously deteriorate to 40.60\% and 37.93\% on CIFAR-100 under identical skewness conditions. Furthermore, contrastive optimization paradigms such as MOON fail to sustain their theoretical advantages, dropping to 40.57\% on CIFAR-100 at $IF=100$. This empirically confirms that traditional parameter aggregation and naive local contrastive pairs become statistically unreliable when a dense label space intersects with severe local sample absence.

\noindent\textbf{2) Resilience and Divergent Vulnerabilities of Specialized Mechanisms.} 
Specialized FL-LT algorithms demonstrate decisively superior resilience compared to traditional baselines, yet our benchmark reveals that their efficacy bifurcates sharply under elevated categorical complexity. Distillation-based frameworks, specifically FedYoYo, maintain a commanding lead, achieving 80.54\% on CIFAR-10 and 52.82\% on CIFAR-100 under the most extreme $IF=100$ configuration. This indicates that localized self-bootstrap feature enhancement serves as a highly scalable defense against double heterogeneity. Similarly, dynamic optimization methodologies including FedLoGe successfully bridge global tail trends with local preferences, securing 44.99\% on CIFAR-100 at $IF=100$. 

Conversely, CReFF plummets to a mere 34.09\% global accuracy on CIFAR-100 at $IF=100$, performing significantly worse than the naive FedAvg baseline (40.60\%). This phenomenon mathematically substantiates that when extreme local skew ($\alpha=0.5$) severely limits minority samples, relying on highly noisy local gradients to optimize virtual features (as in CReFF) becomes fundamentally unstable. 

\subsection{Fine-grained Analysis on Tail Classes (3-Shot Evaluation)}
Relying exclusively on global accuracy frequently masks the true utility of models on rare categories, especially as the label space expands. To uncover the detailed diagnostic capabilities of these algorithms, we employ a fine-grained Head/Middle/Tail stratified evaluation protocol across CIFAR-10, CIFAR-100, and Tiny-ImageNet datasets. The stratified results expose a striking contrast in algorithmic behaviors, confirming the following critical insights:

\noindent\textbf{1) Catastrophic Abandonment in Traditional Baselines.} 
Traditional federated learning methodologies maintain deceptively acceptable overall scores by systematically sacrificing tail classes. While this phenomenon is observable on CIFAR-10, it escalates into a catastrophic collapse on dense datasets like CIFAR-100 and Tiny-ImageNet. As the Imbalance Factor reaches 100, the Tail accuracy for FedProx and FedAvg plummets dramatically, often approaching near-zero performance on the 200-class Tiny-ImageNet despite retaining moderate Head accuracy. This empirically proves that standard optimization paradigms mathematically abandon minority classes to minimize the empirical risk dominated by majority samples, rendering them entirely unsuitable for fairness-critical edge intelligence.

\noindent\textbf{2) Exceptional Tail Recovery via Relational Distillation.} 
Specialized FL-LT algorithms provide a decisive advantage in mitigating tail-class starvation. Specifically, distillation-based frameworks including FedYoYo and FedIC achieve exceptional Tail-class recovery across all three complexity scales. By exploiting localized augmented self-distillation (FedYoYo) or calibrated prototype relations (FedIC), these methods successfully sustain robust tail utility under extreme double heterogeneity without relying on fragile server-side generative mechanisms. Their consistent performance on Tiny-ImageNet validates distillation as a highly stable paradigm for preserving minority-class representations in vast label spaces.

\begin{table*}[htbp]
  \centering
  \footnotesize
  \setlength{\tabcolsep}{4pt}
  \caption{Fine-grained 3-Shot Accuracy Breakdown (Head / Middle / Tail) on CIFAR-10/100 with $\alpha=0.5$.}
  \label{tab:3shot_analysis}
  \begin{tabular}{llcccccc}
    \toprule
    \multirow{2}{*}{Category} & \multirow{2}{*}{Method} & \multicolumn{3}{c}{CIFAR10-LT} & \multicolumn{3}{c}{CIFAR100-LT} \\
    \cmidrule(lr){3-5} \cmidrule(lr){6-8}
    & & IF=10 & IF=50 & IF=100 & IF=10 & IF=50 & IF=100 \\
    & & Head Middle Tail & Head Middle Tail & Head Middle Tail & Head Middle Tail & Head Middle Tail & Head Middle Tail \\
    \midrule
    \multirow{4}{*}{\shortstack[l]{Traditional \\ Baselines}}
    & FedAvg   & 92.4\quad82.2\quad84.8 & 87.7\quad64.9\quad57.3 & 85.2\quad58.5\quad34.7 & 73.8\quad55.5\quad39.2 & 65.2\quad33.5\quad15.6 & 61.2\quad28.2\quad6.1 \\
    & FedProx  & 90.0\quad79.0\quad79.5 & 84.3\quad60.0\quad44.4 & 83.0\quad56.3\quad30.1 & 70.3\quad51.8\quad34.1 & 61.9\quad31.5\quad11.7 & 58.5\quad26.1\quad4.2 \\
    & FedLC    & 92.0\quad82.8\quad86.5 & 86.1\quad63.4\quad59.5 & 86.0\quad59.9\quad39.9 & 73.9\quad54.7\quad36.9 & 64.9\quad34.9\quad14.1 & 60.7\quad28.5\quad7.0 \\
    & MOON     & 91.2\quad81.8\quad85.5 & 86.1\quad65.7\quad57.0 & 84.8\quad64.3\quad39.2 & 72.3\quad54.5\quad37.5 & 65.2\quad33.9\quad14.7 & 61.8\quad26.9\quad5.7 \\
    \midrule
    \multirow{9}{*}{\shortstack[l]{Specialized \\ FL-LT Algorithms}}
    & CReFF    & 83.2\quad77.6\quad85.0 & 82.3\quad68.3\quad59.4 & 87.5\quad60.3\quad53.8 & 56.8\quad45.7\quad31.9 & 49.5\quad30.6\quad18.1 & 47.6\quad27.7\quad10.1 \\
    & CLIP2FL  & 85.6\quad83.3\quad89.5 & 81.8\quad71.7\quad80.5 & 81.2\quad71.3\quad74.6 & 62.7\quad50.4\quad38.6 & 57.3\quad35.3\quad15.3 & 53.6\quad32.0\quad8.8 \\
    & CCVR     & 92.1\quad82.0\quad86.6 & 87.4\quad67.5\quad65.1 & 86.0\quad64.1\quad52.3 & 74.2\quad57.1\quad41.2 & 64.5\quad36.3\quad21.4 & 61.7\quad32.9\quad10.8 \\
    & RUCR     & 88.7\quad80.6\quad85.1 & 84.8\quad66.2\quad52.3 & 83.6\quad58.0\quad35.6 & 70.5\quad53.3\quad36.0 & 63.3\quad33.2\quad13.6 & 60.3\quad27.1\quad7.8 \\
    & FedETF   & 88.4\quad82.4\quad85.9 & 85.5\quad71.7\quad68.5 & 83.1\quad67.8\quad57.3 & 62.8\quad51.8\quad37.1 & 56.0\quad33.9\quad14.7 & 51.9\quad26.1\quad6.7 \\
    & FedLoGe  & 91.4\quad87.0\quad86.1 & 92.5\quad73.1\quad66.7 & 94.3\quad71.8\quad60.5 & 80.6\quad58.7\quad48.2 & 81.3\quad50.5\quad19.6 & 79.8\quad48.9\quad15.1 \\
    & FedIC    & 73.8\quad69.8\quad79.8 & 67.9\quad71.3\quad83.3 & 70.9\quad70.1\quad76.3 & 43.5\quad41.7\quad36.5 & 36.7\quad35.8\quad26.9 & 35.2\quad34.0\quad23.2 \\
    & FedGraB  & 98.7\quad85.6\quad84.8 & 98.7\quad79.8\quad65.1 & 98.2\quad72.6\quad57.6 & 79.0\quad58.9\quad43.1 & 69.9\quad36.5\quad16.0 & 64.9\quad27.9\quad8.3 \\
    & FedYoYo  & 88.9\quad84.0\quad83.4 & 84.1\quad80.1\quad78.3 & 82.0\quad78.8\quad75.0 & 78.7\quad64.4\quad55.0 & 72.5\quad47.2\quad35.6 & 67.5\quad45.7\quad26.4 \\
    \bottomrule
  \end{tabular}
\end{table*}

\begin{table}[t]
    \centering
    \caption{Comparison of 3-shot classification accuracy on TinyImageNet-LT under the setting of IF=50 and $\alpha=0.5$.}
    \label{tab:tinyimagenet}
    \setlength{\tabcolsep}{6pt}
    \begin{tabular}{lcccc}
        \toprule
        \multirow{2}{*}{\textbf{Method}}
        & \multicolumn{4}{c}{\textbf{TinyImageNet-LT}} \\
        \cmidrule(lr){2-5}
        & \textbf{All} & \textbf{Head} & \textbf{Middle} & \textbf{Tail} \\
        \midrule
        FedLoge  & 0.4197 & 0.7035 & 0.3712 & 0.1687 \\
        FedGrab  & 0.4031 & 0.6092 & 0.2543 & 0.1110 \\
        FedYoYo  & 0.3755 & 0.4740 & 0.2913 & 0.2635 \\
        CCVR     & 0.3574 & 0.5076 & 0.2280 & 0.1760 \\
        RUCR     & 0.3272 & 0.4774 & 0.2050 & 0.1335 \\
        FedETF   & 0.2957 & 0.4422 & 0.1813 & 0.1010 \\
        CLIP2FL  & 0.2765 & 0.3970 & 0.1880 & 0.1150 \\
        CREFF    & 0.2326 & 0.3066 & 0.1587 & 0.1585 \\
        FedIC    & 0.2197 & 0.2518 & 0.1883 & 0.2645 \\
        \bottomrule
    \end{tabular}
\end{table}

\begin{figure*}[htbp]
  \centering
  \begin{minipage}{0.32\textwidth}
    \centering
    \includegraphics[width=\linewidth]{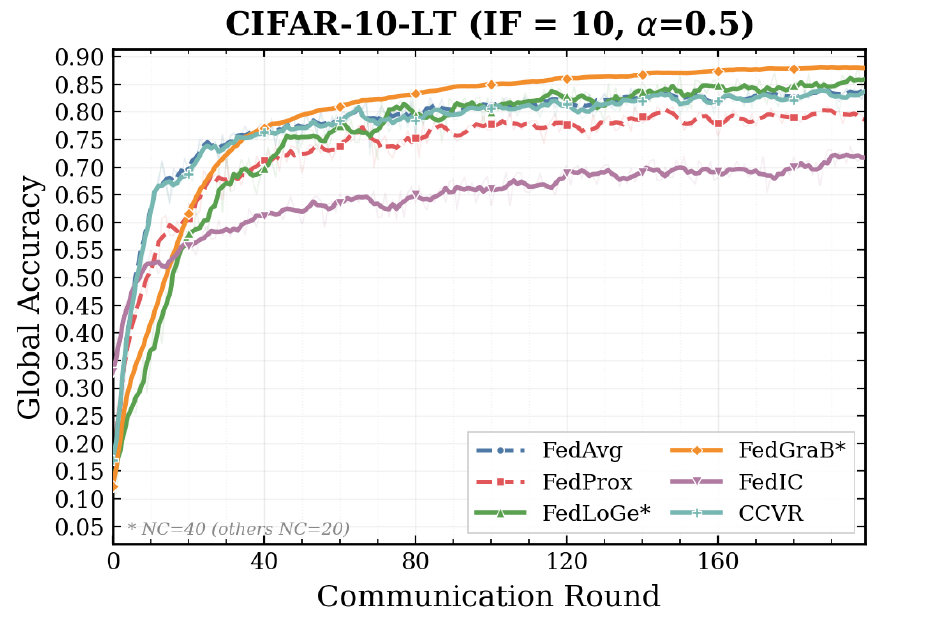}
    \centerline{(a) IF = 10}
  \end{minipage}\hfill
  \begin{minipage}{0.32\textwidth}
    \centering
    \includegraphics[width=\linewidth]{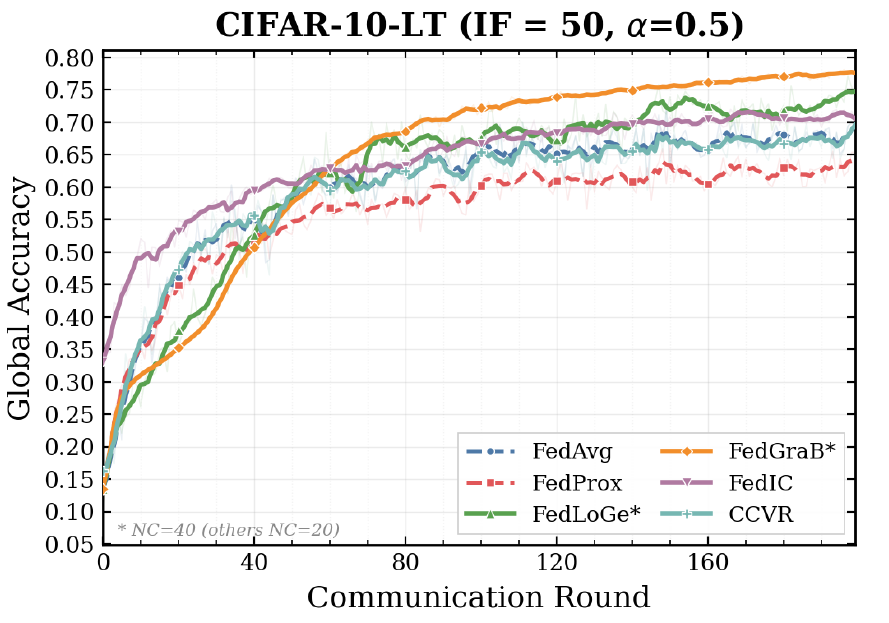}
    \centerline{(b) IF = 50}
  \end{minipage}\hfill
  \begin{minipage}{0.32\textwidth}
    \centering
    \includegraphics[width=\linewidth]{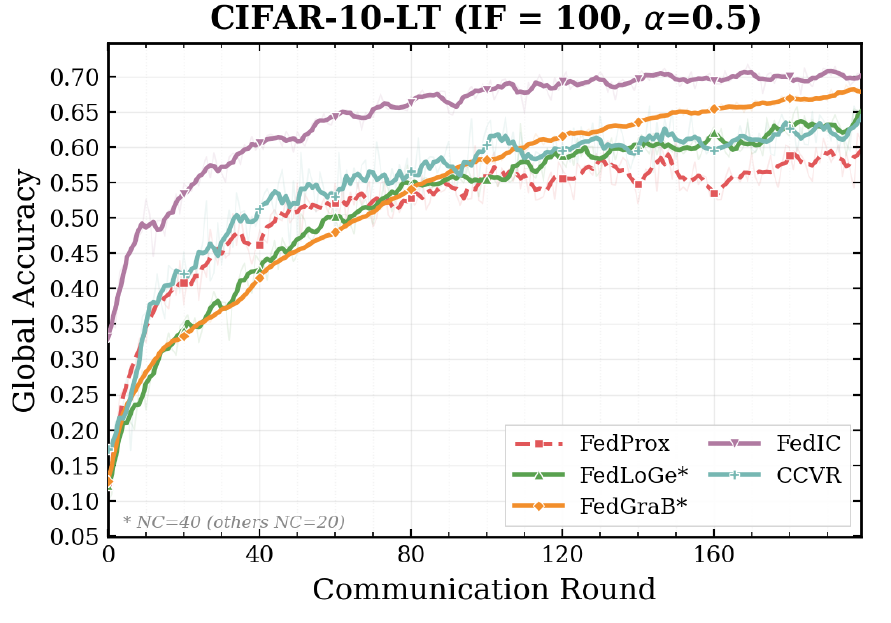}
    \centerline{(c) IF = 100}
  \end{minipage}
  \caption{Convergence curves of global test accuracy across communication rounds under varying Imbalance Factors (IF) with $\alpha=0.5$. Traditional baselines are denoted by dashed lines, while specialized FL-LT algorithms use solid lines. The pale shaded regions represent the raw accuracy fluctuations before moving average smoothing.}
  \label{fig:convergence}
\end{figure*}

\subsection{Convergence and Stability Analysis}
To visualize the underlying learning dynamics and optimization stability, we utilize CIFAR-10 as a representative case study. As illustrated in Figure  \ref{fig:convergence} various IF levels.

\textbf{1) Stability Degradation in Baselines:} While baselines converge smoothly at IF=10 (Figure \ref{fig:convergence}a), they exhibit severe oscillation and early stagnation as skewness intensifies (IF=100, Figure \ref{fig:convergence}c). FedProx struggles to climb above the 60\% threshold, visualizing how traditional aggregation becomes trapped in suboptimal local minima when tail gradients are continuously overwritten.

\textbf{2) Robust Convergence of FL-LT:} In contrast, specialized algorithms demonstrate both faster initial ascent and higher asymptotic stability. Specifically, FedIC exhibits an exceptionally rapid ascent in the initial 20--40 communication rounds across all IF settings. Methods incorporating feature-level calibration, including CCVR and FedIC, maintain smooth learning trajectories even under extreme conditions. This proves that customized interventions effectively stabilize gradient updates throughout the training lifecycle, preventing the minority classes from being overwhelmed.

\begin{figure*}[htbp]
  \centering
  \includegraphics[width=0.85\textwidth]{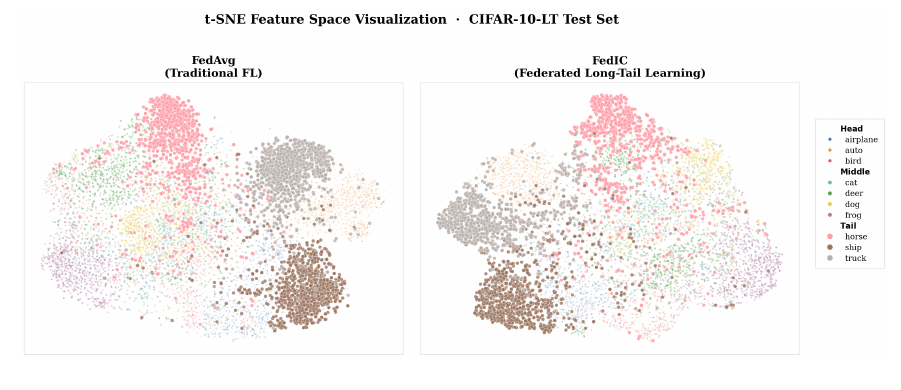}
  \caption{t-SNE visualization of the global model's feature space on the CIFAR-10-LT test set. (Left) Traditional FL (FedAvg) exhibits severe feature entanglement, where tail classes, such as horse, ship, and truck, collapse into the feature clusters of majority classes. (Right) FedIC maintains highly structured representations with clear decision boundaries, effectively preventing minority classes from being absorbed by the head classes.}
  \label{fig:tsne}
\end{figure*}

\subsection{Feature Space Visualization (t-SNE)}
To intuitively explain the performance disparity, we project the feature space into 2D via t-SNE. Given the severe visualization entanglement and overlapping inherent in dense label spaces like CIFAR-100 or Tiny-ImageNet, we provide the projection of the CIFAR-10-LT test set as a lucid diagnostic example (Figure \ref{fig:tsne}). 

As illustrated in the left panel, FedAvg consistently fails to learn discriminative features for minority samples. Representations of tail classes (horse, ship, truck) are heavily entangled with majority clusters, confirming the Feature Collapse phenomenon where rare category representations are overwritten. 

Conversely, the right panel shows that FedIC yields a highly structured and separable feature space. It enforces intra-class compactness and inter-class separability, allowing tail classes to form independent clusters. This provides compelling visual evidence for the superior Few-shot tail accuracy observed in Table \ref{tab:3shot_analysis}.

\subsection{System Efficiency and Resource Trade-offs}
Finally, we address the SOTA Resource Paradox by profiling hardware-critical metrics (Table \ref{tab:system_efficiency}).

\begin{table}[htbp]
  \centering
  \caption{System Efficiency Analysis: Theoretical Communication Payload per round and Empirical Peak Memory Footprint (IF=50). $|\mathbf{w}|$ denotes model size, $C$ is class count, $d$ is feature dimension, and $|P|$ is prototype size.}
  \label{tab:system_efficiency}
  \resizebox{0.48\textwidth}{!}{
  \begin{tabular}{llcc}
    \toprule
    \multirow{2}{*}{\textbf{Method}} & \textbf{Theoretical} & \multicolumn{2}{c}{\textbf{Empirical Memory Footprint}} \\
    \cmidrule(lr){3-4}
    & \textbf{Comm. Payload} & \textbf{CPU Memory} & \textbf{GPU VRAM} \\
    \midrule
    FedAvg & $\mathcal{O}(|\mathbf{w}|)$ & 117.26 MB & 290.83 MB \\
    FedProx & $\mathcal{O}(|\mathbf{w}|)$ & 117.26 MB & 328.44 MB \\
    MOON & $\mathcal{O}(|\mathbf{w}|)$ & 117.26 MB & 427.16 MB \\
    FedLC & $\mathcal{O}(|\mathbf{w}| + C)$ & 117.26 MB & 328.49 MB \\
    \midrule
    CReFF & $\mathcal{O}(|\mathbf{w}| + |P|)$ & 117.26 MB & 100.72 MB \\
    \textbf{FedETF} & $\mathcal{O}(|\mathbf{w}|)$ & 117.26 MB & \textbf{85.86 MB} \\
    CCVR & $\mathcal{O}(|\mathbf{w}| + C\cdot d)$ & 117.26 MB & 295.55 MB \\
    RUCR & $\mathcal{O}(|\mathbf{w}|)$ & 117.26 MB & 422.72 MB \\
    CLIP2FL & $\mathcal{O}(|\mathbf{w}| + \text{VLM})$ & 117.26 MB & 463.05 MB \\
    FedIC & $\mathcal{O}(|\mathbf{w}| + |P|)$ & 276.50 MB & 229.97 MB \\
    FedYoYo & $\mathcal{O}(|\mathbf{w}| + |P|)$ & 117.26 MB & 234.74 MB \\
    \midrule
    \rowcolor{red!10} FedLoGe & $\mathcal{O}(|\mathbf{w}| + C)$ & 117.26 MB & \textbf{1.77 GB} \\
    \rowcolor{red!10} FedGraB & $\mathcal{O}(|\mathbf{w}| + \text{Grad})$ & 117.26 MB & \textbf{2.61 GB} \\
    \bottomrule
  \end{tabular}
  }
\end{table}

The profiling exposes a stark Pareto trade-off often neglected in literature:

\textbf{1) The Hidden Resource Trap:} While accuracy leaders like FedGraB and FedLoGe perform well, they incur massive VRAM footprints (2.61 GB and 1.77 GB respectively) due to intensive gradient buffers and optimization bridging. This renders them impractical for standard edge devices.

\textbf{2) Resilience of Geometric Priors:} In contrast, FedETF achieves remarkable efficiency (85.86 MB VRAM). By fixing the classifier geometry, it bypasses gradient computation for the classification head, demonstrating the superior Pareto trade-off for resource-constrained deployment.

\textbf{3) Practical Equilibrium:} FedYoYo demonstrates a fascinating equilibrium, achieving superior tail recovery with a modest 234.74 MB VRAM footprint, emerging as a practical choice for real-world FL-LT scenarios.

\section{Conclusion and Future Directions}

\subsection{Conclusion}
In this paper, we introduced FedLTLib, a comprehensive, scalable, and highly reproducible benchmark framework specifically designed for Federated Long-Tailed Learning (FL-LT). By mathematically decoupling and parameterizing the double heterogeneity, which represents the superposition of global class imbalance and local non-IID partitions, FedLTLib addresses the critical lack of standardized evaluation protocols in current FL research. 

Through an extensive empirical evaluation of 13 representative algorithms across systematically varied heterogeneity settings, we exposed the fundamental vulnerabilities of traditional FL paradigms. Our fine-grained 3-Shot evaluations and t-SNE visual diagnostics demonstrated that standard aggregation mechanisms, such as FedAvg and FedProx, achieve misleadingly high overall accuracy by severely sacrificing tail classes. Conversely, specialized FL-LT algorithms validated the necessity of customized long-tail interventions. Paradigms leveraging personalized global-local bridging, like FedLoGe, or feature-level calibration, including CCVR and FedIC, exhibited remarkable robustness.

\subsection{Future Directions}
Based on the empirical insights derived from our benchmark, we identify promising trajectories for future research in Federated Long-Tailed Learning:

\textbf{1. Lightweight and Resource-Efficient FL-LT:} Our system efficiency evaluations revealed a stark accuracy-efficiency trade-off. High-performing algorithms often incur substantial computational or communication overheads, such as excessive gradient balancing computations or the transmission of massive prototype sets, which are inherently antagonistic to the resource-constrained nature of edge devices. Future research must prioritize lightweight interventions, such as Parameter-Efficient Fine-Tuning (PEFT) \cite{bian2025survey} tailored for long-tailed distributions, or communication-free logit adjustments.

\textbf{2. Privacy-Preserving Tail Enhancement:} Several state-of-the-art FL-LT methods, particularly Virtual Feature Retraining paradigms, rely on sharing class prototypes, feature covariance matrices, or knowledge distillation logits with the central server. While effective for tail calibration, these mechanisms inadvertently expand the attack surface. Developing mathematically rigorous ``Differentially Private Federated Learning for Long-Tail'' algorithms that can enhance tail representation without leaking sensitive feature-level information remains an open challenge.

\textbf{3. Leveraging Foundation Models for the Tail:} The emergence of Visual-Language Models (VLMs) like CLIP \cite{radford2021learning} and large-scale foundation models presents a paradigm shift \cite{yuan2022decentralized}. As hinted by the robustness of CLIP-based features, demonstrated by methods like CLIP2FL [27] in our benchmark, leveraging zero-shot generalization via Federated Prompt Tuning \cite{che2023federated} or efficient adapter mechanisms \cite{jia2022visual} to adapt foundation models to extremely skewed local data will be a transformative direction for next-generation FL-LT.

\bibliographystyle{IEEEtran}
\bibliography{mybib}

\vfill
\end{document}